\documentclass{article} 

\usepackage{etoolbox}
\usepackage{xcolor}
\usepackage{hyperref}
\usepackage{caption}

\newtoggle{arxiv}
\toggletrue{arxiv} 

\iftoggle{arxiv}{
    \usepackage[
        a4paper,
        inner=25mm,
        outer=25mm,
        top=25mm,
        bottom=25mm
    ]{geometry}

    \usepackage[T1]{fontenc}
    \usepackage[tt=false,type1=true]{libertine}
    \usepackage[libertine]{newtxmath}

    \usepackage[round,authoryear]{natbib}

    \hypersetup{
        colorlinks=true,
        linkcolor=blue,
        citecolor=blue,
        urlcolor=blue
    }

    \setlength{\parskip}{6pt}
    \setlength{\parindent}{0pt}

    \usepackage{fancyhdr}
    \pagestyle{fancy}
    \fancyhf{}

    \fancyhead[L]{IsoAlign}
    \fancyhead[R]{Niousha Sadjadi, et al. (2026)}
    \fancyfoot[C]{\thepage}

    \captionsetup{
        font=small,
        labelfont=it
    }

    \usepackage{authblk}

    \setlength{\affilsep}{1em}

}{
    \usepackage{iclr2027_conference,times}
}

\usepackage{amsmath,amsfonts,bm}
\usepackage{wrapfig}

\def\eqref#1{equation~\ref{#1}}

\def\1{\bm{1}}

\DeclareMathAlphabet{\mathsfit}{\encodingdefault}{\sfdefault}{m}{sl}
\SetMathAlphabet{\mathsfit}{bold}{\encodingdefault}{\sfdefault}{bx}{n}

\usepackage{url}
\usepackage{booktabs}       
\usepackage{amsfonts}       
\usepackage{nicefrac}       
\usepackage{microtype}      
\usepackage{color,colortbl} 
\usepackage{wrapfig,lipsum,booktabs}
\usepackage{enumitem}
\usepackage{svg}
\usepackage{subcaption}
\usepackage{amsmath}
\usepackage{cleveref}
\usepackage{multirow}
\usepackage{placeins}
\usepackage{float}

\newcommand{\IsoAlign}{{\textsc{IsoAlign} }}

\title{IsoAlign: Isoform-Aware DNA-Protein Alignment via Joint-Embedding Predictive Architectures}

\iftoggle{arxiv}{
\author[1,2,*]{Niousha Sadjadi}
\author[2]{Arman Seyed-Ahmadi}
\author[3]{Tony Tu}
\author[4]{Micaela Elisa Consens}
\author[3]{Kerr Ding}
\author[2]{Alif Munim}
\author[1]{Lila Kari}
\author[5]{Zehui Li}
\author[2,4]{Barry B. Rubin}
\author[4,6]{Bo Wang}
\author[2,6,*]{Vallijah Subasri}

\affil[ ]{%
    \mbox{\textsuperscript{1}University of Waterloo}\quad
    \mbox{\textsuperscript{2}Peter Munk Cardiac Centre, University Health Network}\quad
    \mbox{\textsuperscript{3}Georgia Institute of Technology}\quad
    \mbox{\textsuperscript{4}University of Toronto}\quad
    \mbox{\textsuperscript{5}Imperial College London}\quad
    \mbox{\textsuperscript{6}Vector Institute}
}
\affil[*]{Corresponding authors: \texttt{nsadjadi@uwaterloo.ca}, \texttt{vallijah.subasri@uhn.ca}}
}
{}

\date{}%
\begin{document}

\maketitle
\begin{abstract}
Genomic foundation models (GFMs) have emerged as a powerful paradigm for learning DNA sequence representations, yet existing self-supervised pretraining strategies overlook alternative splicing and underperform compared to specialized models. To address this gap, we introduce IsoAlign, a multimodal framework that grounds genomic pretraining in isoform-aware DNA-protein alignment to learn biologically meaningful representations. IsoAlign aligns transcripts to their corresponding protein isoforms, capturing both coding sequences and broader regulatory contexts, using two Joint-Embedding Predictive Architectures (JEPA): DNA-Protein JEPA (DP-JEPA), which predicts protein representations from DNA using contrastive alignment, and Isoform-Masked JEPA (IM-JEPA), which models relationships among alternative isoforms of the same gene through masked protein prediction. We evaluate the learned DNA representations of these architectures across a multi-metric evaluation suite spanning tasks, including the prediction of pathogenicity, fitness, gene expression, and TF binding. Our results demonstrate that isoform-aware cross-modal alignment using JEPA significantly improves functional representations and downstream performance, establishing a promising direction for GFM pretraining.
\end{abstract}

\section{Introduction}
\label{sec:intro}

Genomic foundation models (GFMs) have emerged as a promising paradigm for learning high-dimensional representations of DNA sequences. Inspired by the success of large language models (LLMs) \citep{brown2020language} and protein foundation models (PFMs) \citep{lin2023evolutionary, rives2021}, GFMs leverage transformer-based architectures pretrained on large genomic corpora to learn the underlying regulatory syntax shaped by evolutionary constraints, enabling a unified feature space for diverse downstream biological tasks \citep{zhou2023dnabert,nguyen2023hyenadna,schiff2024caduceus,dalla2025nucleotide}.

However, current pretraining strategies for GFMs, which primarily leverage masked language modeling (MLM) and next-token prediction (NTP), fail to deliver consistent or meaningful performance gains over specialist models \citep{marin2023bend}. In fact, previous studies report that randomly initialized models perform on par with, or even outperform, pretrained GFMs after fine-tuning \citep{schreiber2024, zeng2023, tang2025evaluating}. This suggests that these standard self-supervised objectives fail to generate representations that adequately capture the biological complexity of genomic data. A key limitation is that they treat genomic sequences as flat token streams, without explicitly modeling the hierarchical and multi-stage processes through which genetic information is expressed.

A central bottleneck in connecting genomic variation to phenotypic function is the diversity introduced by alternative splicing.  Although translation of a given coding sequence into its corresponding amino-acid sequence is deterministic under the genetic code, the relationship between a genomic locus and its protein products is not necessarily one-to-one. In eukaryotes, alternative splicing can generate multiple coding transcripts from the same gene, producing protein isoforms with distinct structural properties, subcellular localizations, and biochemical interaction partners \citep{kelemen2013}. In humans, many genes express multiple transcript isoforms, with some loci producing dozens of distinct proteoforms \citep{aebersold2018}. Thus, the challenge is not to rediscover the codon mapping itself, but to learn representations that capture biologically meaningful information shared across nucleotide and protein sequences despite the diversity of transcript and protein products associated with genomic loci.

Despite this biological centrality, alternative splicing remains a blind spot for existing GFMs. Because most models operate on fixed-length genomic windows or gene-level aggregates, they inherently overlook isoform-specific regulation. Consequently, current architectures struggle to resolve how subtle sequence variations propagate through the splicing machinery to alter functional outcomes. This highlights a critical architectural gap: while GFMs learn representations of genomic sequences and PFMs capture properties of protein sequences, the molecular correspondence between coding transcripts and their protein isoforms remains underexplored.

In this work, we propose \textsc{IsoAlign}, a multimodal framework designed to learn isoform-aware genomic representations through cross-modal latent alignment. Rather than relying on token reconstruction losses and scaling alone, our approach utilizes transcript-isoform-aligned DNA-protein pairs and \textbf{Joint-Embedding Predictive Architectures (JEPA)} to explicitly model the functional consequences of genomic variation. By aligning DNA and protein representations that capture high-level biological properties beyond the deterministic codon-to-amino-acid mapping, we propose a pretraining strategy to move GFMs beyond pattern recognition toward functional discovery and precise variant interpretation.

\paragraph{Major Contributions} 
\begin{enumerate}
\item \textbf{\textsc{IsoAlign} Framework:} We introduce a model-agnostic paradigm that grounds genomic pretraining in multimodal biological alignment rather than flat sequence reconstruction. By semantically mapping transcript representations to their corresponding protein isoforms, \textsc{IsoAlign} explicitly captures the functional diversity introduced by alternative splicing across coding sequences and broader regulatory contexts.

\item \textbf{Isoform-Aware Predictive Architectures (JEPA):} We propose two complementary architectures: \textbf{DP-JEPA}, which predicts protein representations from DNA anchors in a shared latent space, and \textbf{IM-JEPA}, which predicts masked protein targets from visible isoform context to capture functional relationships among splice variants.

\item \textbf{Cross-Modal Objective Benchmarking:} Through architecture-matched controls, we quantify the impact of cross-modal alignment, demonstrating that JEPAs learn more biologically meaningful representations than conventional Masked Language Modeling (MLM) and Contrastive Learning (CL).

\item \textbf{Comprehensive Biological Evaluation:} We establish a  benchmark spanning general and splicing-sensitive tasks, evaluating representational robustness in fitness prediction, splicing variant effect prediction, 
and tissue-specific gene expression.

\end{enumerate}


\section{Preliminaries and Related Work}
\label{sec:related_work}
\paragraph{Limitations of Genomic Foundation Models.} While scaling laws have dominated the development of PFMs and GFMs \citep{avsec2025alphagenome, zhou2023dnabert}, recent evidence suggests that increasing parameter counts does not monotonically improve performance \citep{hou2025understanding, li2024feature}. Moreover, large models impose prohibitive training and inference costs, which limits accessibility and iteration. Self-supervised objectives based on token reconstruction, such as Masked Language Modeling (MLM), often bias GFMs toward local sequence statistics at the expense of global functional constraints \citep{ye2025predicting}. Long-context capabilities have been positioned as a solution for capturing distal genomic dependencies; however, recent studies suggest that current models struggle to effectively exploit this expanded context \citep{yang2026extending, cheng2025dnalongbench}. These findings highlight a broader limitation: naive DNA-only self-supervised pretraining lacks the grounding necessary to map genomic variation to its functional downstream consequences, often resulting in representations that lack clinical relevance and sensitivity to genomic heterogeneity \citep{lu2025genomic}.

\paragraph{Moving Beyond Traditional Genomic Pretraining.} Current pretraining strategies yield marginal improvements over randomly initialized baselines, with performance frequently influenced by tokenizer choice rather than the pretraining objective itself \citep{vishniakov2024genomic, bairethinking}. These limitations have motivated a shift toward objectives that encourage learned representations to capture biologically meaningful features beyond local sequence statistics \citep{consens2026predicting, ye2025predicting}. Moreover, training data composition plays a critical role in representation quality. The SPACE framework \citep{zhang2024space} demonstrates that pretraining on functionally annotated profiles improves generalization through a species-aware mixture-of-experts architecture. Contrastive learning (CL) has also emerged as a strong alternative to masking-based objectives. Orthrus leverages decoupled contrastive learning (DCL) with biologically motivated augmentations for RNA modeling \citep{fradkin2026orthrus}, mirroring gains observed in protein sequence–structure alignment \citep{zhang2023structure} and protein–DNA binding prediction \citep{chen2025dpac}. More recently, joint-embedding predictive architectures (JEPA), including JEPA-DNA \citep{larey2026jepa} and GenoJEPA \citep{wang2026nucleotides}, move beyond local nucleotide reconstruction toward latent-space prediction, encouraging representations grounded in functional semantics. However, despite these architectural advancements, overall performance gains remain modest, and these models largely fail to demonstrate meaningful improvements on complex, clinically relevant downstream tasks such as variant effect prediction (VEP). Collectively, these developments reflect a broader paradigm shift  from conventional sequence-only masked objectives toward biologically informed and latent-aware formulations designed to learn richer and more transferable representations \citep{he2025generalized, consens2026predicting}.

\paragraph{Multimodal Integration and Biological Alignment.} Biological systems are inherently multimodal, operating across a hierarchical continuum of DNA (regulatory), RNA (transcriptional), and protein (functional) layers. Recent foundation models aim to unify these sequences into a shared latent space through diverse architectural strategies. ProTrek integrates protein language models with natural language and structural tokens \citep{su2025trimodal}, while LucaOne trains a single transformer jointly on nucleotide and amino acid sequences to achieve cross-modal alignment \citep{he2025generalized}. To bridge these modalities beyond simple translation, models like CDBridge utilize adaptive token-merging and tissue-specific conditioning to map genomic sequences to expression levels \citep{yu2026cdbridge}, and BioLangFusion aligns embeddings at the codon level for precise property prediction \citep{mollaysa2025biolangfusion}. Complementing these approaches, RESM-2 leverages cross-modal distillation to transfer evolutionary knowledge from ProtESM-2 to RNA by mapping sequences into pseudo-protein representations, enabling protein-informed RNA modeling \citep{zhang2025resm}. 
Despite these advances, a critical bottleneck persists: many architectures still treat genomic sequences as static inputs or rely on coarse gene-level associations, overlooking how genetic variation propagates through alternative splicing \citep{mitschka2022context} and isoform-specific regulation to contribute to disease \citep{scotti2016rna}.

\section{Dataset} 
\label{sec:dataset}
We construct DNA-protein training pairs from the Broad's Gene-to-Protein (G2P) database \citep{kwon2024genomics}, linking Ensembl transcripts to UniProt protein isoforms. For each gene, we extract genomic DNA (retaining intron/exon/UTR annotations), transcript regions (with CDS boundaries), and all associated protein isoforms. Pairs are generated via two alignment strategies outlined in Appendix \ref{app:training-data}. We train each of our architectures using both alignment strategies and report the best-performing configuration.

\section{Method}
\label{sec:methods}
  \subsection{Double Encoder Framework}
  
  Our architecture employs a dual-encoder framework to jointly model DNA and protein sequences in a shared embedding space. Two modality-specific encoders generate fixed-dimensional representations, which are aligned through projection heads for cross-modal learning.

  \paragraph{DNA Encoder (X-Encoder).} The DNA encoder $f_\theta$ is a trainable transformer-based model that processes tokenized DNA sequences. Given an input sequence $\mathbf{x} \in
  \mathcal{V}_{\text{DNA}}^L$, it produces token-level hidden states $\mathbf{H}_x = f_\theta(\mathbf{x}) \in \mathbb{R}^{L \times D_x}$. We evaluate two backbone architectures: \textbf{NTv2-500m}~\citep{dalla2025nucleotide}, a transformer pretrained on multi-species genomic data, and \textbf{AlphaGenome} \citep{avsec2025alphagenome}, a CNN-based model that operates on one-hot encoded sequences and produces chunk-level embeddings.

  \paragraph{Protein Encoder (Y-Encoder).} The protein encoder $g_\phi$ processes protein sequences into fixed representations. We employ a pretrained protein language model, \textbf{ESM2-150M}~\citep{lin2023evolutionary}, trained on millions of protein sequences with masked language modeling. Given a protein sequence $\mathbf{y}$, the encoder produces token-level hidden states $\mathbf{H}y = g_\phi(\mathbf{y}) \in \mathbb{R}^{M \times D_y}$. By default, the protein encoder is frozen ($\phi$ fixed) to provide stable prediction targets.

\paragraph{Pooling and Projection.} Both encoders produce variable-length token representations that are aggregated via multi-head attention pooling with a learnable query vector:
  \begin{equation}
      \mathbf{z} = \text{MultiHeadAttn}(\mathbf{q}, \mathbf{H}, \mathbf{H}),
  \end{equation}
  where $\mathbf{q} \in \mathbb{R}^{D}$ is a learnable query, 
  $\mathbf{H} \in \mathbb{R}^{L \times D}$ are token representations,
  and we use 8 attention heads.
  Pooled representations are passed through 3-layer MLP projection heads with
  batch normalization and ReLU activations:
  \begin{equation}
      \mathbf{z}_x = \pi_x(\text{pool}(\mathbf{H}_x)) \in \mathbb{R}^{512}, \quad
      \mathbf{z}_y = \pi_y(\text{pool}(\mathbf{H}_y)) \in \mathbb{R}^{512},
  \end{equation}
  with hidden dimension 256. Embeddings are $\ell_2$-normalized before similarity computation.

\subsection{IsoAlign JEPA Architectures}
\subsubsection{DNA-Protein JEPA (DP-JEPA)}
\paragraph{Architecture.} We introduce DNA-Protein Joint-Embedding Predictive Architecture (DP-JEPA), a cross-modal self-supervised framework that learns to predict protein representations from DNA sequences in a shared latent space, outlined in Figure \ref{fig:jepa}a. It consists of: (1) a trainable DNA encoder $f_\theta$, (2) a protein encoder $g_\phi$ (frozen by default), and (3) a TransformerPredictor $h_\psi$ that predicts protein embeddings from DNA representations.

The TransformerPredictor is a stack of transformer encoder layers (default: 4 layers, 8 heads) that processes the full sequence of DNA hidden states with attention masking, then mean-pools across valid positions:
  \begin{equation}
      \hat{\mathbf{z}}_y = \pi_{\text{pred}}\bigl(\mathrm{LN}(\mathrm{MeanPool}(h_\psi(\mathbf{H}_x, \mathbf{m}_x)))\bigr),
  \end{equation}
where $\mathbf{m}_x$ is the attention mask excluding padding tokens.

The target protein embedding applies stop-gradient to the encoder output while keeping the projection trainable:
  \begin{equation}
      \mathbf{z}_y = \pi_y\bigl(\mathrm{sg}(\mathrm{pool}(g_\phi(\mathbf{y})))\bigr).
  \end{equation}

  Both $\pi_{\text{pred}}$ and $\pi_y$ are multi-layer projection heads mapping to a shared embedding space.

\paragraph{Training Objective.} DP-JEPA employs contrastive alignment in the shared projection space. For a batch of genes, each potentially having multiple protein isoforms, we compute the loss per isoform slot and average:
  \begin{equation}
      \mathcal{L}_{\text{DP-JEPA}} = \frac{1}{|\mathcal{V}|}\sum_{j \in \mathcal{V}}
      \mathcal{L}_{\text{contrast}}(\hat{\mathbf{Z}}_y, \mathbf{Z}_{y_j}; \tau),
  \end{equation} where $\mathcal{V}$ denotes valid isoform slots (non-padded entries with $\geq 2$ samples for contrastive discrimination). The target projection $\pi_y$ is frozen under stop-gradient to provide stable targets.

\subsubsection{Isoform-Masked JEPA (IM-JEPA)}

\begin{figure}[ht]
    \centering

    \begin{subfigure}{\linewidth}
        \centering
        \includegraphics[width=0.9\linewidth]{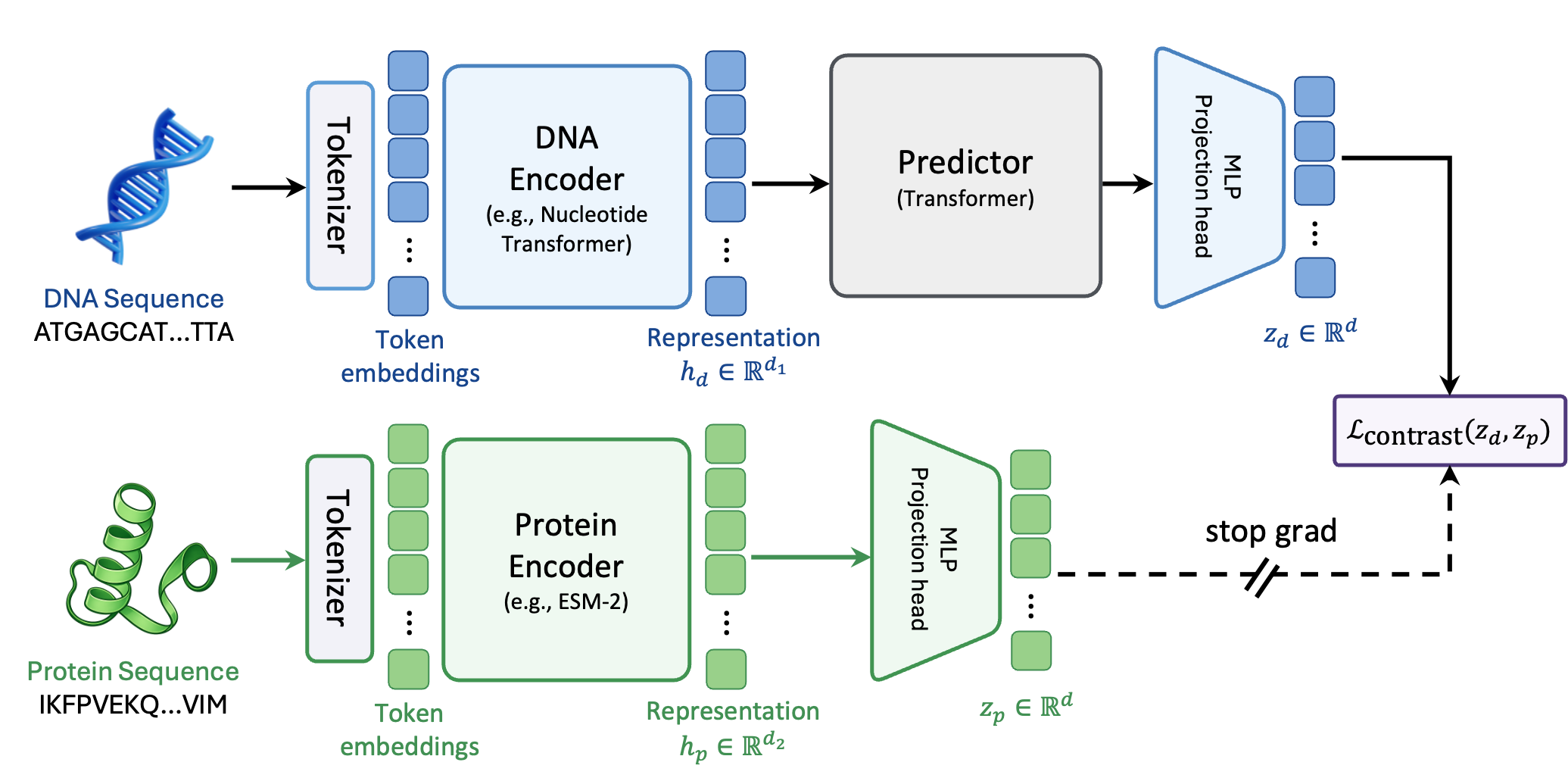}
        \caption{\textbf{DP-JEPA (DNA-Protein Joint-Embedding Predictive Architecture):} A dual-encoder framework where DNA sequences are processed by a trainable DNA encoder (e.g., NTv2) and projected through a Transformer predictor to align with protein representations from a frozen protein encoder (e.g., ESM2). The model learns cross-modal alignment via contrastive loss in a shared latent space.}
        \label{fig:sub1}
    \end{subfigure}

    \vspace{0.2cm}

    \begin{subfigure}{1.0\linewidth}
        \centering
        \includegraphics[width=\linewidth]{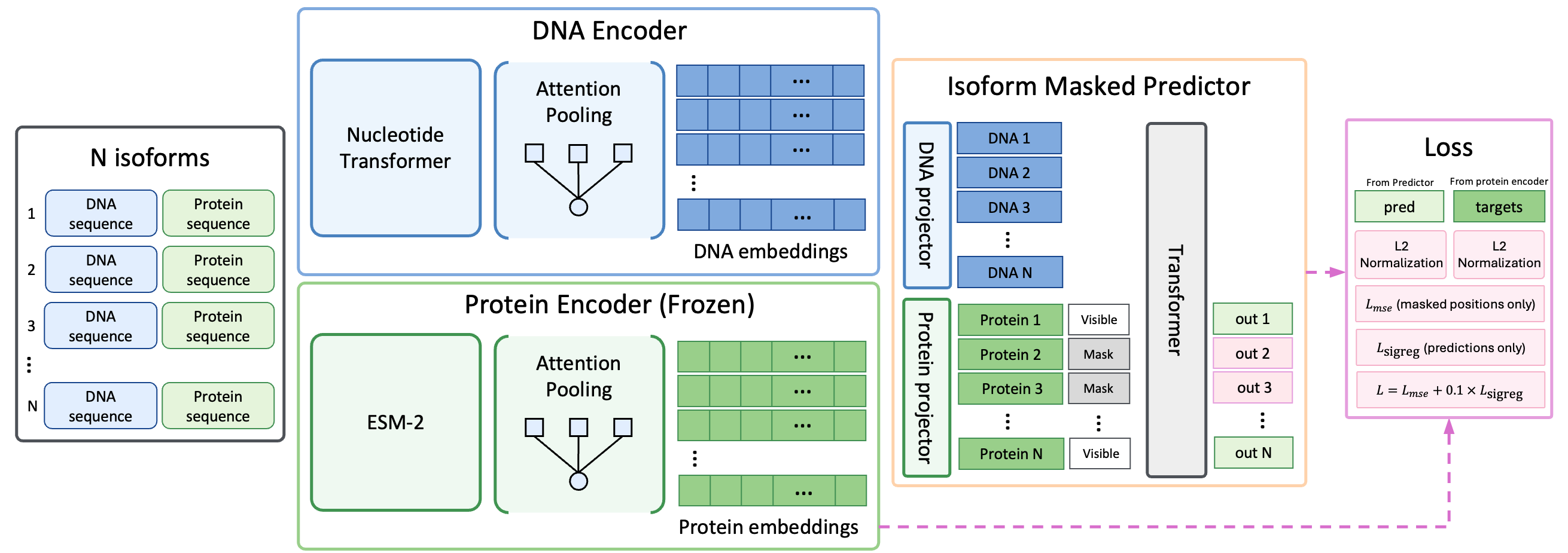}
        \caption{\textbf{IM-JEPA (Isoform-Masked JEPA):} An extension of DP-JEPA that jointly models multiple transcript-isoform pairs from the same genes. DNA sequences for all isoforms are encoded and pooled, while protein isoforms are randomly masked ($\sim$50\%). The Transformer Predictor uses DNA anchors as identity queries to predict masked protein embeddings, with $\ell_2$-normalized MSE loss computed only on masked positions and SIGReg regularization to prevent collapse.}
        \label{fig:sub2}
    \end{subfigure}

    \caption{\textbf{\textsc{IsoAlign} JEPA Architectures}}
    \label{fig:jepa}
\end{figure}

\paragraph{Architecture.} While DP-JEPA aligns each transcript-isoform pairs through cross-modal prediction, it does not explicitly model relationships among alternative isoforms of the same gene. To incorporate this intra-gene structure, we introduce Isoform-Masked JEPA (IM-JEPA), which jointly processes multiple transcript--isoform pairs and predicts masked protein isoforms from their DNA anchors and the remaining visible isoform context. Specifically, for genes with multiple transcript-isoform pairs
    $\{(\mathbf{x}_i, \mathbf{y}_i)\}_{i=1}^N$, we extend DP-JEPA with object-level masking
    inspired by Causal-JEPA~\citep{nam2026causal}, outlined in Figure \ref{fig:jepa}b. Approximately 50\% of isoforms are randomly
    masked per sample while all DNA anchors remain visible, ensuring at least one isoform
    stays visible for context. The predictor operates on \emph{pooled} encoder representations
    and employs three types of learnable embeddings: (1) modality embeddings distinguishing
    DNA from protein isoforms, (2) slot embeddings indexing each pair, and
    (3) a learnable mask token $\mathbf{m}$. For masked isoforms, we inject \emph{identity
    queries} derived from their DNA anchors via a learned projection $\rho$:
    \begin{equation}
        \mathbf{s}_i^{\text{iso}} = \begin{cases}
            \pi_{\text{iso}}(\mathbf{z}_{\mathbf{y}_i}) + \mathbf{e}_{\text{iso}} + \mathbf{e}_i & \text{if visible} \\
            \mathbf{m} + \rho(\pi_{\text{dna}}(\mathbf{z}_{\mathbf{x}_i})) + \mathbf{e}_{\text{iso}} + \mathbf{e}_i & \text{if masked}
        \end{cases},
    \end{equation}
    where $\mathbf{z}_{\mathbf{x}_i} = \text{pool}(f_\theta(\mathbf{x}_i))$ and
    $\mathbf{z}_{\mathbf{y}_i} = \text{pool}(g_\phi(\mathbf{y}_i))$ are pooled encoder outputs,
    and $\pi_{\text{dna}}$, $\pi_{\text{iso}}$ are input projections with layer normalization.

    The full input sequence concatenates all DNA representations with isoform
    representations, processed by a bidirectional Transformer encoder:
    \begin{equation}
        \hat{\mathbf{Z}} = h_\psi\bigl([\pi_{\text{dna}}(\mathbf{z}_{\mathbf{x}}^{(1:N)}) + \mathbf{e}_{\text{dna}} + \mathbf{e}_{\text{slot}};\; \mathbf{s}^{(1:N)}]\bigr).
    \end{equation}

\paragraph{Training Objective and Regularization.} For IM-JEPA, the loss is computed on masked and valid (non-padded) positions using $\ell_2$-normalized embeddings:
    \begin{equation}
        \mathcal{L}_{\text{IM-JEPA}} = \frac{1}{|\mathcal{M} \cap \mathcal{V}|} \sum_{i \in \mathcal{M} \cap \mathcal{V}} \bigl\|\bar{\hat{\mathbf{z}}}_i - \bar{\mathbf{z}}_i\bigr\|_2^2 +
  \lambda_{\text{reg}}
     \mathcal{L}_{\text{SIGReg}},
    \end{equation}
    where $\bar{\mathbf{z}} = \mathbf{z} / \|\mathbf{z}\|_2$, $\mathcal{M}$ denotes masked positions,
    $\mathcal{V}$ denotes valid (non-padded) pairs, and $\mathcal{L}_{\text{SIGReg}}$ is an anti-collapse
regularization term.


Specifically, SIGReg \citep{balestriero2025lejepa} encourages predictions to follow an isotropic Gaussian $\mathcal{N}(0, \mathbf{I})$ via characteristic function matching. Predictions are projected onto $K$ random unit vectors $\{\mathbf{p}_k\}_{k=1}^{K}$ (resampled each iteration) and compared to the Gaussian characteristic function $\varphi(t) = e^{-t^2/2}$:                     
\begin{equation}
      \mathcal{L}_{\text{SIGReg}} = \frac{1}{K}\sum_{k=1}^{K} \int_0^{t_{\max}} w(t) \left[                                                                    
      \bigl(\bar{c}_k(t) - e^{-t^2/2}\bigr)^2 + \bar{s}_k(t)^2 \right] dt,                                    
\end{equation}
where $\bar{c}_k(t) = \mathbb{E}[\cos(t \cdot \mathbf{p}_k^\top \hat{\mathbf{z}})]$ and $\bar{s}_k(t) = \mathbb{E}[\sin(t \cdot \mathbf{p}_k^\top \hat{\mathbf{z}})]$ are empirical characteristic function components, and $w(t) \propto e^{-t^2/2}$ weights errors by target magnitude. We use $K{=}1024$ projections with trapezoidal integration over $t \in [0, 3]$.

\subsection{Architecture-Matched Pretraining Controls}
To isolate the performance gains of our JEPA objectives from the underlying dual-encoder architecture and increased parameter count, we implement several architecture-matched controls. These controls utilize the exact same DNA encoder, frozen protein encoder, and projection heads, but replace the JEPA objective with MLM or CL:

  \paragraph{Masked Language Modeling (MLM).}
  We implement four MLM configurations with a 15\% masking rate. Cross-modal conditioning utilizes either \textbf{sequence-level} (global pooling) or \textbf{token-level} (position-specific) multi-head
  cross-attention (8 heads), with linear projections to align hidden dimensions between modalities.
  \begin{enumerate}                                         
      \item \textbf{Standard Joint:} Independent MLM with combined loss:  $\mathcal{L}_{\mathrm{Joint}} = \lambda_D \mathcal{L}_{\mathrm{MLM}}^D + \lambda_P \mathcal{L}_{\mathrm{MLM}}^P$.
 
      \item \textbf{P$\rightarrow$D:} Unmasked protein predicts masked DNA:
      $\hat{\mathbf{H}}_D \!=\! \mathrm{LN}(\mathrm{Attn}(\mathbf{H}_D, \mathbf{W}_P\mathbf{H}_P) \!+\! \mathbf{H}_D)$.
      \item \textbf{D$\rightarrow$P:} Unmasked DNA predicts masked protein:
      $\hat{\mathbf{H}}_P \!=\! \mathrm{LN}(\mathrm{Attn}(\mathbf{H}_P, \mathbf{W}_D\mathbf{H}_D) \!+\! \mathbf{H}_P)$.
      \item \textbf{Bidirectional:} Both modalities are masked; combines (2) and (3) for mutually enhanced reconstruction.
  \end{enumerate}
  
\paragraph{Contrastive Learning.}
We align DNA-protein embeddings using three loss variants, governed by a biological correctness matrix $\mathbf{C} \in \{0, 1\}^{N \times N}$. All variants utilize a 8,192-entry momentum memory bank ($m=0.999$) \citep{he2020moco}.

\begin{itemize}
    \item \textbf{InfoNCE:} Maximizes matched-pair similarity using learnable temperature $\tau$ (init 0.5).
    \item \textbf{DCL:} Decoupled loss~\citep{yeh2022decoupled} separating positive and negative terms for independent weighting: $\mathcal{L}_{\text{DCL}} = \lambda_{\text{pos}}(-\bar{s}_{\text{pos}}) + \lambda_{\text{neg}}\log\sum\exp(s_{\text{neg}})$.
    \item \textbf{CLIP:} Symmetric bidirectional NCE with low initial temperature ($\tau_0=0.1$, learnable) for sharp distribution alignment \cite{radford2021learning}.
\end{itemize}

\section{Results}


\subsection{Deep Mutation Scan Fitness Prediction} 
\begin{figure}[t]
    \centering

        \includegraphics[width=\linewidth]{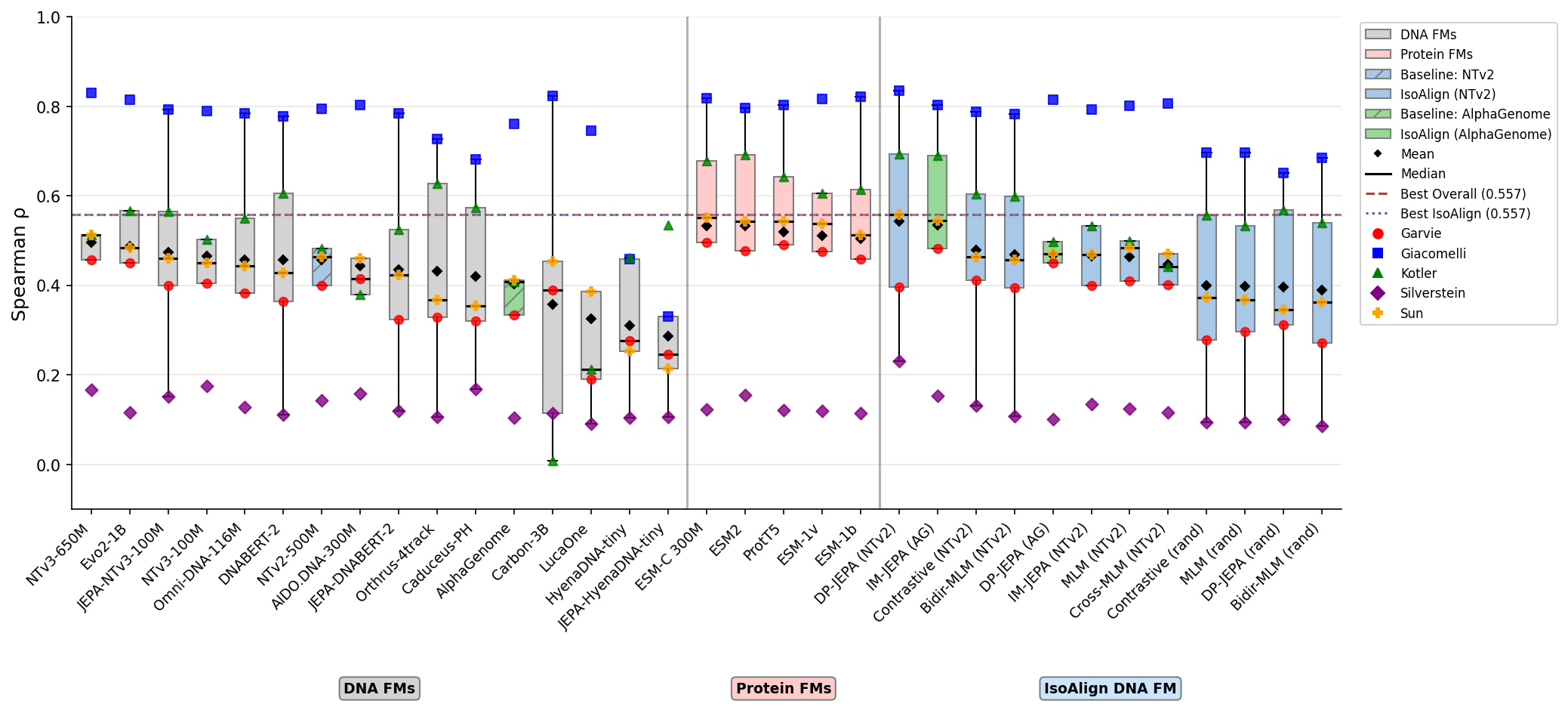}
        \caption{\textbf{DMS Fitness Prediction.} Spearman correlation between embedding-based variant scores and fitness scores on Deep Mutational Scanning (DMS) datasets. Models are grouped by category: DNA FMs (gray), Protein FMs (red), and \textsc{IsoAlign} models (blue). \textsc{IsoAlign} consistently outperforms unimodal DNA baselines, demonstrating that cross-modal pretraining with protein-derived inductive biases improves variant effect prediction without task-specific fine-tuning.}



    \label{fig:dms}
\end{figure}

We evaluated linear probing for fitness prediction on five DMS datasets from four human genes—\textit{CBS}, \textit{P53}, \textit{PDE3A}, and \textit{GDI1}—following~\citet{brixi2026genome} (Appendix~\ref{app:dms}). Each dataset contains wild-type and mutant sequences with experimentally measured fitness scores. GFMs and \textsc{IsoAlign} are evaluated using DNA representations, while PFMs are evaluated separately using protein representations for comparison. We use group $K$-fold cross-validation stratified by amino acid mutation identity to prevent codon leakage, where synonymous codons encoding identical substitutions could otherwise appear across splits and artificially inflate GFM performance. We evaluate \textsc{IsoAlign} variants combining different GFMs (NTv2/AlphaGenome) with ESM2 under multiple pretraining objectives and initialization schemes. The best-performing models use our JEPA objectives (Figure~\ref{fig:dms}), achieving median Spearman correlations of $0.557 \pm 0.21$ (DP-JEPA) and $0.554 \pm 0.22$ (IM-JEPA). Notably, DP-JEPA performs better with NTv2, while IM-JEPA favors AlphaGenome, and both substantially outperform their respective GFM baselines (DP-JEPA: +0.094 over NTv2; IM-JEPA: +0.136 over AlphaGenome) as well as architecture-matched cross-modal MLM and contrastive controls, demonstrating that isoform-aware alignment improves DNA representations. Across 16 GFMs and 5 PFMs, our best \textsc{IsoAlign} model outperforms all baselines. Per-model setup and regression protocol are detailed in Appendix~\ref{app:impl}.
    
\subsection{Probing DNA Representations for Splicing Variant Effects}

We evaluate variant effect prediction (VEP) from DNA representations, with the goal of distinguishing pathogenic variants (those that disrupt normal gene function) from benign ones. We focus on two splicing-related single-nucleotide variant (SNV) datasets: ClinVar Splicing, which contains clinically annotated splice-site variants, and MFASS (Massively Functional Assay of Splicing SNVs), a functional assay of splicing disruption~\citep{chong2019multiplexed}; dataset details are provided in Appendix~\ref{app:vep}. For each variant, we compare the wild-type and mutant representations, quantify the resulting perturbation using \texttt{cosine} and $\ell_2$ distances, and measure performance using AUPRC under leave-one-chromosome-out (LOCO) cross-validation. Additional representation-based evaluations are reported in Appendix~\ref{app:vep_result}, and per-model setup and evaluation protocols are detailed in Appendix~\ref{app:impl}.

As shown in Table~\ref{tab:zero-shot-eval}, cross-modal DNA-protein pretraining can increase the functional discriminability of the underlying DNA representation, although the magnitude of the improvement depends on the backbone and training objective. 
On ClinVar Splicing, the large-scale baselines Evo2 and Carbon achieve strong performance, consistent with their larger model capacity and extensive pretraining. Despite its smaller backbone, DP-JEPA performs best among our objectives, substantially improving NTv2-500M from 0.32 to 0.63 \texttt{cosine} AUPRC and AlphaGenome from 0.61 to 0.63. In contrast, existing JEPA-DNA adaptations do not consistently improve their parent DNA models and can even degrade performance, indicating that gains are not inherent to JEPA-based pretraining alone. 
MFASS presents a substantially more challenging setting, with only 1,050 pathogenic variants among 28,972  (3.6\%), and correspondingly near-chance AUPRC values across the baseline models, including the larger models. Nevertheless, DP-JEPA produces substantial gains over its baseline representations: with NTv2-500M, $\ell_2$ AUPRC increases from 0.04 to 0.17, while with AlphaGenome, \texttt{cosine} AUPRC increases from 0.29 to 0.36. IM-JEPA provides little improvement over NTv2-500M and  underperforms the parent AlphaGenome representation. Overall, DP-JEPA consistently improves or maintains the functional sensitivity of its underlying DNA representations across these settings, whereas the effectiveness of IM-JEPA is more backbone-dependent.

\begin{table*}[t]
  \centering
  \caption{Variant effect prediction from DNA representations, evaluated by AUPRC using two metrics. Values in parentheses indicate changes relative to the parent DNA model; \underline{underlined} values indicate the best performance per column, and \textbf{bold} values are the best \textsc{IsoAlign} model per backbone.}
  \label{tab:zero-shot-eval}

  \small
  \renewcommand{\arraystretch}{0.85}
  \setlength{\tabcolsep}{5pt}

  \begin{tabular}{llllll}
      \toprule
      & & \multicolumn{2}{c}{ClinVar Splicing}
      & \multicolumn{2}{c}{MFASS} \\
      \cmidrule(lr){3-4} \cmidrule(lr){5-6}
      Model & Params & \texttt{Cosine} & $\ell_2$ & \texttt{Cosine} & $\ell_2$ \\
      \midrule

      DNABERT-2~\citep{zhou2023dnabert}
      & 117M & 0.14 & 0.15 & 0.04 & 0.04 \\
      JEPA-DNABERT-2~\citep{larey2026jepa}
      & 117M & 0.14 & 0.15 & 0.04 & 0.04 \\
      
      HyenaDNA-tiny~\citep{nguyen2023hyenadna}
      & 0.44M & 0.16 & 0.15 & 0.04 & 0.04 \\
      JEPA-HyenaDNA-tiny
      & 0.44M & 0.15 & 0.16 & 0.04 & 0.04 \\

      Caduceus-PH~\citep{schiff2024caduceus}
      & 7.7M & 0.20 & 0.18 & 0.04 & 0.04 \\

      LucaOne~\citep{he2025generalized}
      & 1.8B & 0.15 & 0.16 & 0.05 & 0.06 \\

      Orthrus-4track~\citep{fradkin2026orthrus}
      & 10M & 0.18 & 0.16 & 0.04 & 0.04 \\
      Orthrus-MLM
      & 10.7M & 0.18 & 0.19 & 0.04 & 0.04 \\

      Omni-DNA-116M~\citep{li2025omni}
      & 116M & 0.15 & 0.15 & 0.04 & 0.04 \\

      AIDO.DNA-300M~\citep{ellington2024accurate}
      & 300M & 0.52 & 0.58 & 0.05 & 0.09 \\

      NTv3-100M~\citep{boshar2025foundational}
      & 100M & 0.51 & 0.66 & 0.10 & 0.12 \\
      JEPA-NTv3-100M
      & 100M & 0.47 & 0.55 & 0.12 & 0.09 \\

      Evo2-1B-base~\citep{brixi2026genome}
      & 1B & 0.38 & 0.85 & 0.05 & 0.16 \\
      Carbon~\citep{allal2026carbon}
      & 3B & \underline{0.63} & \underline{0.90} & 0.05 & 0.08 \\

      \midrule

      NTv2-500M~\citep{dalla2025nucleotide}
      & 500M & 0.32 & 0.73 & 0.04 & 0.04 \\
      + DP-JEPA (CLIP)
      & 520M & \textbf{0.63} {\scriptsize (+.31)} & 0.67  & 0.08 {\scriptsize (+.04)} & \textbf{0.17} {\scriptsize (+.13)} \\
      + DP-JEPA (DCL)
      & 520M & 0.39 {\scriptsize (+.07)} & 0.47  & 0.05 {\scriptsize (+.01)} & 0.06 {\scriptsize (+.02)} \\
      + DP-JEPA (InfoNCE)
      & 520M & 0.58 {\scriptsize (+.26)} & 0.55   & \textbf{0.07} {\scriptsize (+.03)} & 0.12 {\scriptsize (+.08)} \\
      + IM-JEPA
      & 520M & 0.35 {\scriptsize (+.03)} & \textbf{0.74} {\scriptsize (+.01)} & 0.04  & 0.05 {\scriptsize (+.01)} \\

      \midrule

      AlphaGenome~\citep{avsec2025alphagenome}
      & 450M & 0.61 & 0.57 & 0.29 & \underline{0.31} \\
      + DP-JEPA (CLIP)
      & 470M & \textbf{0.63} {\scriptsize (+.02)} & \textbf{0.60} {\scriptsize (+.03)} & \textbf{\underline{0.36}} {\scriptsize (+.07)} & 0.31 \\
      + IM-JEPA
      & 470M & 0.33 & 0.30 & 0.06 & 0.09 \\

      \bottomrule
  \end{tabular}
\end{table*}



  
\subsection{Supervised Finetuning}

We evaluate our model on two primary biological tasks: i) TF-DNA binding specificity (TFLink) ~\citep{liska2022tflink} and ii) tissue-specific transcript expression (GTEx), using two fine-tuning strategies: i) Linear Probing and ii) Parameter Efficient Fine-Tuning (LoRA), described in Appendix~\ref{app:sectionc}. Per-baseline fine-tuning configurations are detailed in Appendix~\ref{app:impl}.

\paragraph{TFLink.} 
We benchmark on the TFLink dataset using a hard negative split with 35\% within-family negatives, specifically designed to assess discrimination between structurally similar TFs with overlapping sequence motifs. Details of the dataset construction, sampling procedure, and TF-family composition are provided in Appendix~\ref{app:TFlink}. As shown in Table \ref{tab:tflink_results}, frozen representations alone underperform, suggesting TF binding specificity requires task adaptation beyond generic sequence embeddings. In contrast, LoRA fine-tuning substantially improves performance, outperforming prior GFM and specialist baselines. These results suggest that isoform-aware representations can capture regulatory sequence features highly predictive of TF-DNA interactions.

\begin{table}[htbp]                                       
    \centering                                            
    \caption{TFLink transcription factor binding prediction (mean $\pm$ SD over 4 training seeds).}
    \label{tab:tflink_results}
    \begin{tabular}{llccc}                                
    \toprule                                              
    \textbf{Model} & \textbf{Training} & \textbf{Params} & \textbf{AUROC} & \textbf{AUPRC} \\                                                    
    \midrule

    LucaOne~\citep{he2025generalized} & LoRA & 1.8B & 0.785 $\pm$ 0.001 & 0.778 $\pm$ 0.002 \\
    BPNet~\citep{avsec2021base} & Full & 0.12M & 0.861 $\pm$ 0.004 & 0.851 $\pm$ 0.003 \\
    ChromBPNet~\citep{pampari2023bias} & Full & 6M & 0.867 $\pm$ 0.005 & 0.851 $\pm$ 0.007 \\
    AlphaGenome~\citep{avsec2025alphagenome} & Full & 450M & 0.892 $\pm$ 0.002 & 0.879 $\pm$ 0.003 \\
    
    \midrule
    \textsc{IsoAlign} & LoRA & 650M & \textbf{0.934 $\pm$ 0.001} & \textbf{0.928 $\pm$ 0.001} \\

    \bottomrule
    \end{tabular}
  \end{table}

\paragraph{GTEx.} 
We use expression data from five tissues in GTEx \citep{lonsdale2013genotype}: Brain, Heart, Kidney, Liver, and Stomach ($N$=445,888 records across 19,587 unique genes). We implement gene-level splitting (8:1:1), ensuring that all transcripts from a given gene remain within the same train, validation, or test partition to prevent data leakage. As shown in Table~\ref{tab:gtex_results}, \textsc{IsoAlign} achieves the best overall performance, outperforming both LucaOne and AlphaGenome across most tissues. \textsc{IsoAlign} attains the highest overall Spearman correlation ($\rho=0.725$) and Pearson correlation ($r=0.738$), with particularly strong gains in Kidney and Liver tissue prediction, demonstrating the benefit of protein-informed genomic representations for tissue-specific expression modeling.

\begin{table}[htbp]                                                                                                                                                                                         
      \centering                                                                                                                                                                                            
      \caption{GTEx tissue-specific expression prediction.}                                                                                                                                
      \label{tab:gtex_results}                                                                                                                                                                                
      \resizebox{\textwidth}{!}{%
      \begin{tabular}{lcccccccccccc}                                                                                                                                                                        
      \toprule
      & \multicolumn{2}{c}{\textbf{Overall}} & \multicolumn{2}{c}{\textbf{Brain}} & \multicolumn{2}{c}{\textbf{Heart}} & \multicolumn{2}{c}{\textbf{Kidney}} & \multicolumn{2}{c}{\textbf{Liver}} &
  \multicolumn{2}{c}{\textbf{Stomach}} \\
      \cmidrule(lr){2-3} \cmidrule(lr){4-5} \cmidrule(lr){6-7} \cmidrule(lr){8-9} \cmidrule(lr){10-11} \cmidrule(lr){12-13}
      \textbf{Model} & $\rho$ & $r$ & $\rho$ & $r$ & $\rho$ & $r$ & $\rho$ & $r$ & $\rho$ & $r$ & $\rho$ & $r$ \\
      \midrule
      LucaOne & 0.432 & 0.430 & 0.448 & 0.437 & 0.456 & 0.455 & 0.383 & 0.380 & 0.426 & 0.417 & 0.453 & 0.453 \\
      AlphaGenome & 0.717 & 0.732 & 0.702 & 0.729 & 0.732 & 0.743 & 0.740 & 0.742 & 0.683 & 0.704 & 0.719 & 0.737 \\
      \midrule
      \textsc{IsoAlign} & \textbf{0.725} & \textbf{0.738} & 0.716 & 0.739 & 0.735 & 0.744 & 0.745 & 0.747 & 0.697 & 0.715 & 0.722 & 0.736 
      \\
      \bottomrule
      \end{tabular}%
      }
  \end{table}

\iftoggle{arxiv}{
\FloatBarrier
}{}

\subsection{Sequence Ablation Analysis}
\label{sec:ablation} 
  \begin{wrapfigure}{r}{0.5\textwidth}
      \vspace{-1em}
      \centering
      \includegraphics[width=0.48\textwidth]{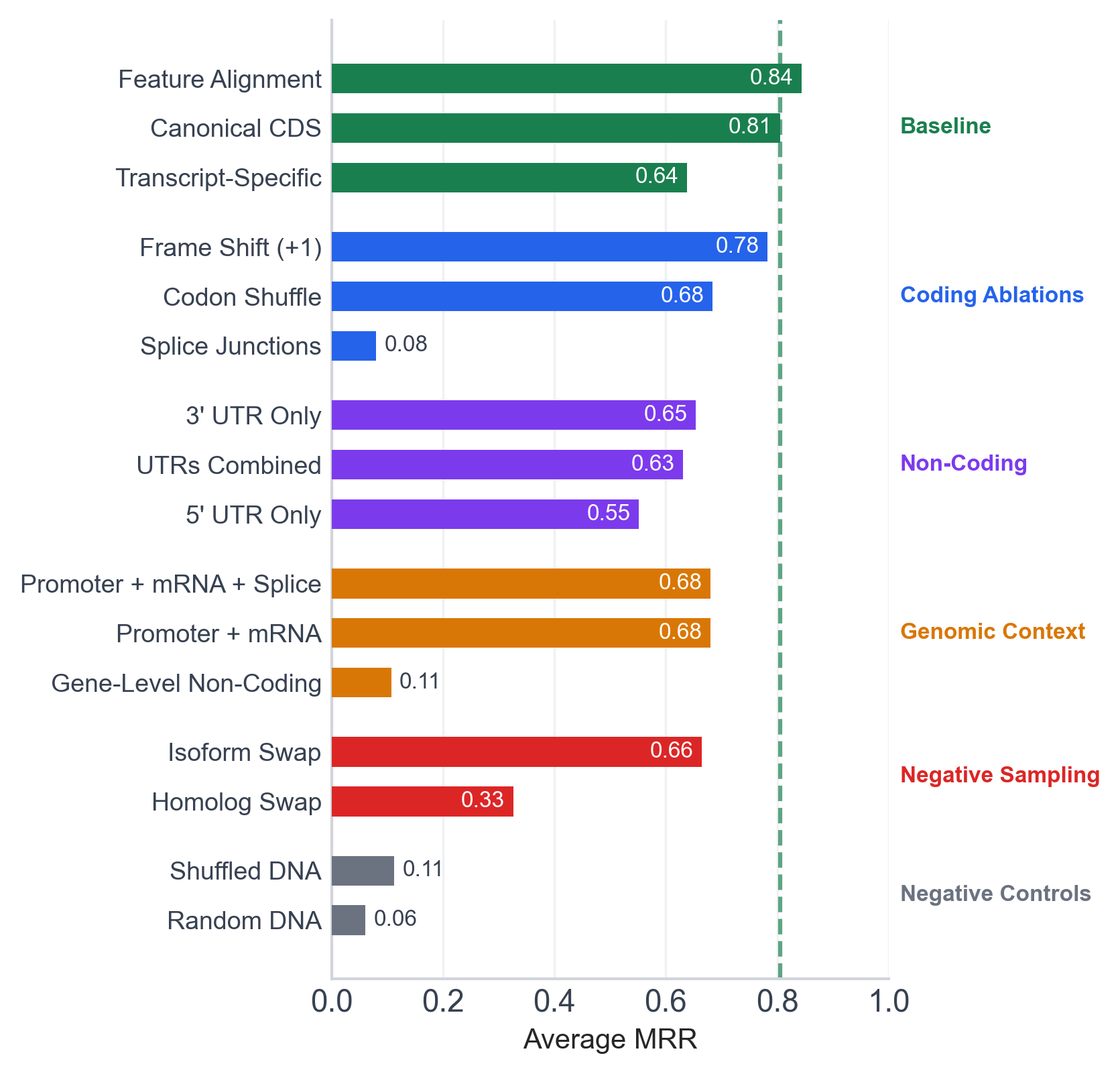}
      \caption{Cross-modal retrieval (Avg MRR) by ablation category. Dashed line: CDS baseline.}
      \label{fig:ablations}
      \vspace{-1em}
  \end{wrapfigure}
To show that  \textsc{IsoAlign}  learns beyond a codon lookup table, we systematically ablated input sequences and evaluated cross-modal retrieval (Figure~\ref{fig:ablations}; Appendix~\ref{app:ablation}). If \textsc{IsoAlign} relied on deterministic codon-to-amino-acid translation, codon-shuffled sequences (which preserve the encoded protein while altering nucleotide positions) should match or exceed the CDS baseline, while frame-shifted sequences (which produce an entirely different amino acid sequence) should fail entirely. We observe the opposite: codon shuffling degrades retrieval by 19\%, whereas frame shifting degrades only 7\%, demonstrating that \textsc{IsoAlign} captures positional codon dependencies, including codon usage bias, translational efficiency signals, and local sequence context, beyond simple translation lookup. Training exclusively on untranslated regions (UTRs), which contain no coding information, retains 65--78\% of CDS-baseline performance far above the random DNA baseline (6\%), confirming that \textsc{IsoAlign} learns regulatory sequence features that correlate with protein function independently of the genetic code. Finally, isoform-swapped negatives (wrong isoform, same gene) retain 79\% of baseline retrieval, whereas homolog-swapped negatives (protein from a related gene family) degrade to 39\%, indicating that \textsc{IsoAlign} distinguishes subtle isoform-level differences while recognizing that homologs, despite sequence similarity, have distinct functional contexts.


\section{Conclusion}
\label{sec:conclusion}
We introduced \textsc{IsoAlign}, a model-agnostic framework for genomic pretraining driven by isoform-aware DNA-protein alignment. By explicitly modeling the functional diversity of alternative splicing via two complementary JEPA architectures (DP-JEPA and IM-JEPA), the framework moves beyond flat sequence reconstruction. 
Our evaluations demonstrate that isoform-aware cross-modal prediction improves functional genomic representations, with DP-JEPA providing consistent gains across fitness and variant effect prediction, while IM-JEPA shows more task- and backbone-dependent improvements. Supervised evaluations further demonstrate strong performance in transcription factor binding and tissue-specific expression prediction. 
Crucially, sequence ablations confirm that these learned representations capture true regulatory syntax rather than merely memorizing deterministic codon mappings. Ultimately, this establishes biologically grounded, cross-modal latent alignment as a highly effective paradigm for GFMs.


\paragraph{Limitations and Future Work.}
\textsc{IsoAlign} is limited by the completeness of DNA-protein databases and curated benchmarks, which may not capture context-specific splicing or real-world patient variation. Future work will evaluate patient-derived cohorts and incorporate epigenomic and structural information. Clinical translation will additionally require safeguards against population bias and privacy vulnerabilities.

\iftoggle{arxiv}{}{
\subsection*{AI use statement}


In this work, we used generative AI tools for assistance with code correction and improvement of the flow and composition of text originally written by the authors, including checking grammar, spelling, and consistency of terminology and notation across sections.
We have not used generative AI tools for the conception of the research, formulation of theoretical arguments or derivations. All AI-assisted code was reviewed and tested by the authors for correctness, and no AI-assisted text was incorporated into the manuscript without being read, edited, and verified by a human co-author. All references were independently checked against their primary sources. We take responsibility for the final content of this work, including text, claims or artifacts produced with the aid of generative AI.

\subsection*{Ethics statement}
This work relies exclusively on publicly available datasets, pretrained models, and computational methods. It does not involve human subjects, personally identifiable information, or private or sensitive data.

\subsection*{Reproducibility statement}
We provide detailed information to support reproducibility of our experiments. 
For the datasets used in our experiments, Appendix~\ref{app:data} provides a detailed description of training and evaluation dataset construction and preprocessing, including DNA-protein pair alignment and dataset-specific filtering procedures. 
For model training and evaluation, Appendix~\ref{app:sectionc} describes the supervised fine-tuning protocols, and Appendix~\ref{app:impl} provides model-specific configurations and evaluation procedures for all benchmarked models. 
For the ablation experiments, Appendix~\ref{app:ablation} describes the construction of each ablated sequence and its evaluation procedure. 
Finally, Appendix~\ref{app:compute} reports the computational resources, training configurations, hyperparameters, and compute requirements used in our experiments.
}




\bibliography{iclr2027_conference}
\bibliographystyle{iclr2027_conference}


\newpage
\section*{Appendix}
\appendix
\section{Dataset Construction and Statistics}  
\label{app:data}
\subsection{Training Data Alignment}
\label{app:training-data}
We construct DNA-protein training pairs using two complementary alignment strategies that differ in the granularity of sequence correspondence.


\paragraph{Transcript-Specific.} Pairs spliced transcripts (exons and UTRs) with CDS-sliced protein subsequences. DNA chunks of 4,608 bp include explicit markers denoting 5'UTR, CDS, and 3'UTR boundaries, with only the corresponding protein region extracted for each chunk.

\paragraph{Feature-Aligned (CDS-Only).} Pairs coding sequences directly with protein counterparts using strict 3:1 codon-to-amino-acid mapping. Exon boundaries are preserved across both modalities to facilitate learning of splice junction patterns. DNA chunks of 3,072 bp are paired with up to 1,024 amino acid tokens.

\subsection{Deep Mutational Scanning (DMS)}  
  \label{app:dms}
  The DMS evaluation dataset measures mutation effects on protein function using experimental fitness scores. Each sample contains a mutated protein sequence, its corresponding DNA sequence, and a fitness score derived from high-throughput functional assays.

  \begin{table}[htbp]
  \centering
  \caption{Deep Mutational Scanning dataset statistics.}
  \label{tab:dms_stats}
  \begin{tabular}{llrrrr}
  \toprule
  Study & Protein & Samples & Protein (aa) & DNA (bp) & Value Range \\
  \midrule
  Kotler et al. 2018 & TP53 & 2,967 & 393 & 1,182 & -4.99--1.51 \\
  Giacomelli et al. 2018 & TP53 & 7,448 & 393 & 1,182 & -4.37--2.84 \\
  Sun et al. 2020 & CBS & 6,757 & 551 & 1,656 & {0.00--1.00} \\
  Garvie et al. 2021 & PDE3A & 7,893 & 1,141 & 3,426 & -3.83--2.47 \\
  Silverstein et al. 2021 & GDI1 & 4,365 & 447 & 1,344 & -7.33--3.09 \\
  \midrule
  \textbf{Total} & & \textbf{29,430} & & & \\
  \bottomrule
  \end{tabular}
  \end{table}

  \subsection{Variant Effect Prediction}
  \label{app:vep}
  \textbf{ClinVar} is a widely-used clinical variant database containing expert-curated pathogenicity labels for human genetic variants ~\cite{landrum2018clinvar}. 
  We focus on ClinVar SNVs within splice-site regions in the immediate vicinity of exon–intron junctions, where conserved splice donor (5') and acceptor (3') motifs regulate RNA splicing. Even single-nucleotide perturbations near these sites can disrupt splicing fidelity and lead to severe functional consequences, including exon skipping or intron retention. Following Avsec et al.~\cite{avsec2025alphagenome}, we define the splice-site region as the interval spanning 6~bp into the intron and 3~bp into the exon at each annotated junction, using \href{https://www.gencodegenes.org/human/stats_46.html}{GENCODE v46} as the transcript reference. We include both donor- and acceptor-proximal variants and restrict the dataset to autosomal and sex-chromosome SNVs with unambiguous ClinVar annotations (Pathogenic/Likely pathogenic or Benign/Likely benign), excluding variants with conflicting or uncertain interpretations. Variants are filtered from the \href{https://ftp.ncbi.nlm.nih.gov/pub/clinvar/vcf_GRCh38/clinvar.vcf.gz}{ClinVar GRCh38 release}, resulting in a dataset containing 7,028 pathogenic and 41,535 benign SNVs.

\textbf{MFASS (Massively Functional Assay of Splicing SNVs)} provides experimentally grounded measurements of splicing disruption, making it a particularly reliable benchmark for VEP. Unlike clinically annotated datasets, MFASS variants were directly assayed using a massively parallel minigene reporter system (SORT-seq)~\cite{chong2019multiplexed}, which quantifies the splicing consequences of thousands of single-nucleotide variants in a controlled cellular context. Variants are labeled as positive if they induce a strong loss-of-function splicing effect, and negative otherwise. Following the benchmark protocol of Avsec et al.~\cite{avsec2025alphagenome}, we retain only SNVs with unambiguous reference and alternate alleles and exclude any sequences containing ambiguous nucleotides.  
The dataset is derived from the publicly available \href{https://github.com/KosuriLab/MFASS/tree/master}{MFASS repository}, enabling a direct evaluation of whether models can detect functionally disruptive splicing variants independent of clinical or population-level annotations.



\subsection{TFLink: TF-DNA Binding Prediction}
\label{app:TFlink}
We construct an evaluation dataset from TFLink \cite{liska2022tflink}, a database of 8.8M experimentally-validated human transcription factor binding sites. The task is binary classification: given a TF protein sequence and a 1024 bp genomic DNA sequence centered on a potential binding site, predict whether binding occurs.

\textbf{Dataset Construction.} Positive examples pair each TF with its validated binding sites. Negative examples pair each TF with binding sites of other TFs---including structurally similar TFs from the same family---creating hard negatives that require the model to learn TF-specific binding preferences rather than generic DNA accessibility. We apply chromosome-stratified sampling with a minimum 10 kb distance between sites to ensure sample independence. The dataset contains 200,000 samples (100,000 positive, 100,000 negative) across 20 TFs from 10 TF families. Table~\ref{tab:tflink_hard_negatives_stratification} summarizes the per-TF statistics.

\begin{table}[htbp]
    \centering
    \caption{TFLink Evaluation Dataset}
    \label{tab:tflink_hard_negatives_stratification}
    \begin{tabular}{llr}
    \toprule
    Stratification & Categories & Samples \\
    \midrule
    \multirow{2}{*}{Binding Status} & Positive & 100,000 \\
     & Negative (same-family) & 100,000 \\
    \midrule
    \multirow{3}{*}{Binding Site Length} & Short ($\leq$12 bp) & 111,982 \\
     & Medium (13--20 bp) & 78,807 \\
     & Long ($>$20 bp) & 9,211 \\
    \midrule
    \multirow{6}{*}{TF Family} & TF\_bZIP & 70,000 \\
     & zf-GATA & 20,000 \\
     & zf-C2H2 & 20,000 \\
     & Fork\_head & 20,000 \\
     & bHLH & 20,000 \\
     & Other (5 families) & 50,000 \\
    \bottomrule
    \end{tabular}
  \end{table}

\section{Additional Variant Effect Prediction Evaluations}
\label{app:vep_result}
To evaluate the predictive power of dna representations, we compare wildtype and mutant embeddings through four representation-based metrics. This includes two \textbf{scalar metrics}: \texttt{cosine} distance at mutated positions and the $\ell_2$ norm of the embedding difference, assessed using AUPRC. Additionally, we employ two \textbf{vector metrics}: element-wise embedding products (\texttt{Mul}) \cite{benegas2025benchmarking} and (\texttt{CLS}) token differences, evaluated through leave-one-chromosome-out (LOCO) cross-validation logistic regression. For models without a \texttt{[CLS]} token, the \texttt{CLS} feature is taken from the final position for autoregressive models and from the first position otherwise.

\begin{table*}[!ht]
  \centering
  \caption{Variant effect prediction evaluation on ClinVar Splicing and MFASS using four metrics. \underline{Underlined} values indicate the best performance per column, and \textbf{bold} values are the best \textsc{IsoAlign} model per backbone.}
  \label{tab:vep-eval-appendix}

  \resizebox{\textwidth}{!}{
  \begin{tabular}{lccccccccc}
      \toprule
      & & \multicolumn{4}{c}{ClinVar Splicing}
      & \multicolumn{4}{c}{MFASS} \\

      \cmidrule(l){3-6} \cmidrule(l){7-10}

      Model & Params
      & \texttt{Cosine} & $\ell_2$ & \texttt{CLS} & \texttt{Mul}
      & \texttt{Cosine} & $\ell_2$ & \texttt{CLS} & \texttt{Mul} \\

      \midrule

      \multicolumn{10}{l}{\textbf{Baselines}} \\

      DNABERT-2~\citep{zhou2023dnabert}
      & 117M & 0.14 & 0.15 & 0.18 & 0.18
      & 0.04 & 0.04 & 0.05 & 0.04 \\
      JEPA-DNABERT-2~\citep{larey2026jepa}
      & 117M & 0.14 & 0.15 & 0.27 & 0.19
      & 0.04 & 0.04 & 0.08 & 0.04 \\

      HyenaDNA~\citep{nguyen2023hyenadna}
      & 6.6M & 0.17 & 0.18 & 0.25 & 0.18
      & 0.04 & 0.04 & 0.05 & 0.04 \\
      HyenaDNA-tiny
      & 0.44M & 0.16 & 0.15 & 0.22 & 0.17
      & 0.04 & 0.04 & 0.04 & 0.04 \\
      JEPA-HyenaDNA-tiny
      & 0.44M & 0.15 & 0.16 & 0.19 & 0.17
      & 0.04 & 0.04 & 0.04 & 0.03 \\
      
      Caduceus-PH~\citep{schiff2024caduceus}
      & 7.7M & 0.20 & 0.18 & 0.19 & 0.19
      & 0.04 & 0.04 & 0.05 & 0.05 \\
      Caduceus-PS
      & 7.7M & 0.20 & 0.45 & 0.42 & 0.42
      & 0.04 & 0.08 & 0.13 & 0.10 \\
      
      LucaOne~\citep{he2025generalized}
      & 1.8B & 0.15 & 0.16 & 0.18 & 0.17
      & 0.05 & 0.06 & 0.09 & 0.07 \\
      LucaOne-gene
      & 1.8B & 0.15 & 0.17 & 0.18 & 0.17
      & 0.05 & 0.06 & 0.08 & 0.06 \\
      
      Orthrus-4track~\citep{fradkin2026orthrus}
      & 10M & 0.18 & 0.16 & 0.14 & 0.16
      & 0.04 & 0.04 & 0.04 & 0.04 \\
      Orthrus-MLM
      & 10.7M & 0.18 & 0.19 & 0.15 & 0.19
      & 0.04 & 0.04 & 0.04 & 0.04 \\

      Omni-DNA-116M~\citep{li2025omni}
      & 116M & 0.15 & 0.15 & 0.55 & 0.21
      & 0.04 & 0.04 & 0.10 & 0.06 \\
      Omni-DNA-300M
      & 300M & 0.15 & 0.14 & 0.49 & 0.21
      & 0.04 & 0.04 & 0.13 & 0.06 \\

      AIDO.DNA-300M~\citep{ellington2024accurate}
      & 300M & 0.52 & 0.58 & 0.52 & 0.52
      & 0.05 & 0.09 & 0.12 & 0.07 \\

      NTv3-650M~\citep{boshar2025foundational}
      & 650M & 0.35 & 0.57 & 0.34 & 0.28
      & 0.04 & 0.07 & 0.08 & 0.07 \\
      NTv3-100M
      & 100M & 0.51 & 0.66 & 0.22 & 0.21
      & 0.10 & 0.12 & 0.07 & 0.08 \\
      JEPA-NTv3-100M
      & 100M & 0.47 & 0.55 & 0.25 & 0.20
      & 0.12 & 0.09 & 0.08 & 0.08 \\

      Evo2-1B-base~\citep{brixi2026genome}
      & 1B & 0.38 & 0.85 & 0.59 & 0.35
      & 0.05 & 0.16 & 0.16 & 0.10 \\
      Evo2-7B-base
      & 7B & 0.57 & 0.88 & \underline{0.72} & 0.36
      & 0.05 & 0.09 & 0.13 & 0.13 \\
      Carbon~\citep{allal2026carbon}
      & 3B & \underline{0.63} & \underline{0.90} & 0.65 & \underline{0.75}
      & 0.05 & 0.08 & 0.09 & 0.06 \\

      \midrule
      
      NTv2-500M~\citep{dalla2025nucleotide}
      & 500M & 0.32 & 0.73 & 0.52 & 0.42
      & 0.04 & 0.04 & 0.05 & 0.04 \\
      + DP-JEPA (CLIP)
      & 520M & \textbf{0.63} & 0.67 & 0.47 & \textbf{0.46}
      & \textbf{0.08} & \textbf{0.17} & \textbf{0.18} & \textbf{0.08} \\
      + DP-JEPA (DCL)
      & 520M & 0.39 & 0.47 & 0.49 & 0.24
      & 0.05 & 0.06 & 0.10 & 0.06 \\
      + DP-JEPA (InfoNCE)
      & 520M & 0.58 & 0.55 & \textbf{0.59} & 0.25
      & 0.07 & 0.12 & \textbf{0.18} & 0.06 \\
      + IM-JEPA
      & 520M & 0.35 & \textbf{0.74} & \textbf{0.59} & \textbf{0.46}
      & 0.04 & 0.05 & 0.07 & 0.04 \\
      \midrule

      AlphaGenome~\citep{avsec2025alphagenome}
      & 450M & 0.61 & 0.57 & 0.63 & 0.30
      & 0.29 & \underline{0.31} & 0.32 & \underline{0.23} \\
      +DP-JEPA (CLIP)
      & 470M & \textbf{0.63} & \textbf{0.60} & 0.61 & 0.29
      & \textbf{\underline{0.36}} & 0.31 & \textbf{\underline{0.36}} & 0.21 \\
      +IM-JEPA
      & 470M & 0.33 & 0.30 & 0.57 & 0.20
      & 0.06 & 0.09 & 0.21 & 0.13 \\

      \midrule
      \multicolumn{10}{l}{\textbf{Architecture-matched pretraining controls (NTv2-500M)}} \\ 
      \midrule


      MLM
      & 520M & 0.27 & 0.73 & 0.59 & 0.39
      & 0.04 & 0.04 & 0.06 & 0.04 \\

      Conditioned MLM
      & 520M & 0.27 & 0.73 & 0.53 & 0.41
      & 0.04 & 0.04 & 0.05 & 0.04 \\

      Bidirectional TCA
      & 520M & 0.50 & 0.61 & 0.63 & 0.49
      & 0.04 & 0.04 & 0.04 & 0.04 \\

      Cross-Modal
      & 520M & 0.52 & 0.48 & 0.42 & 0.43
      & 0.04 & 0.05 & 0.05 & 0.05 \\

      Contrastive (CLIP)
      & 520M & 0.40 & 0.70 & 0.47 & 0.50
      & 0.04 & 0.07 & 0.06 & 0.06 \\

      Contrastive (DCL)
      & 520M & 0.47 & 0.64 & 0.39 & \textbf{0.57}
      & 0.04 & 0.05 & 0.05 & 0.05 \\

      Contrastive (InfoNCE)
      & 520M & 0.39 & 0.69 & 0.46 & 0.52
      & 0.05 & 0.09 & 0.06 & 0.05 \\





      \bottomrule
  \end{tabular}
  }
\end{table*}

\section{Supervised Finetuning}
\label{app:sectionc}

\paragraph{Linear Probing} 
Both the DNA encoder and Protein encoder are frozen. Features are extracted via mean pooling and concatenated into a $d=1664$ representation. 
\paragraph{Parameter-Efficient Fine-Tuning} 
Low-Rank Adaptation (LoRA) was applied to the DNA encoder's query and value attention layers ($r=4$). This configuration introduces $\sim1.4$M trainable adapter parameters while maintaining a frozen Protein encoder, totaling $\sim2.4$M trainable parameters including the head.

\paragraph{Prediction Head} 
For both tasks, we utilize a standardized MLP head:
\begin{equation}
    f(\mathbf{h}) = \mathbf{W}_3 \phi \left( \mathbf{W}_2 \phi \left( \mathbf{W}_1 \text{LN}(\mathbf{h}) \right) \right)
\end{equation}
where $\text{LN}$ is Layer Normalization and $\phi$ denotes GELU activation followed by Dropout ($p=0.2$). For GTEx regression, a 64-dimensional tissue embedding is concatenated to the pooled features before the MLP. Ultra-long sequences are managed via 6-mer tokenization with a 512-token truncation cutoff.

\section{Implementation Details for Benchmarked Models}
\label{app:impl}

This section catalogues per-model setup decisions that diverge from a naïve
Hugging Face load, plus the scoring conventions used across DMS, VEP,
TFLink, and GTEx. Every divergence below is forced by an architectural
constraint of the model in question (context cap, tokenizer choice, missing
head, attention kernel), not by a tunable hyperparameter.

\subsection{Per-Model Setup}

\paragraph{AIDO.DNA-300M~\citep{ellington2024accurate}.} Loaded via the
\texttt{modelgenerator} package (the model's reference RNABert backbone is
not exposed through the standard \texttt{AutoModel} routing). DMS and all
VEP tracks use the model's native context $L{=}3998$
(\texttt{max\_pos\_embeddings}{=}4000); because RNABert uses eager attention
and materialises the full $[B, h, L, L]$ probability tensor, we run with
batch size 8.

\paragraph{Caduceus-PH / Caduceus-PS~\citep{schiff2024caduceus}.}
Bidirectional Mamba SSMs requiring a separate environment
(\texttt{mamba\_ssm} 2.2.4, \texttt{causal\_conv1d} 1.5.0). Single-nt
character tokenizer with no \texttt{[CLS]} token; pooled representations
are mean-pooled across positions.

\paragraph{ChromBPnet~\citep{pampari2023bias}.} The reference TensorFlow/Keras
stack pins old versions incompatible with our PyTorch environment, so we load
weights through \texttt{bpnet-lite}, a PyTorch port that reproduces the Keras
\texttt{.h5} forward bit-equivalently. The K562-pretrained checkpoint
underperformed by $\sim$0.10 AUROC across all fine-tuning recipes we tried;
we report the same architecture trained from scratch.

\paragraph{BPNet~\citep{avsec2021base}.} A 64-filter, 8-layer dilated CNN
trained from scratch. We follow the
original architecture (motif kernel~21, dilated residual kernels~3 with
dilations $2^1,\dots,2^8$) and pair the DNA trunk with the same frozen
ESM2-150M protein encoder and MLP head used for ChromBPnet. The
\textsc{BPNet (Full)} row matches the small-CNN supervised baseline of
\citet{avsec2021base}.

\paragraph{DNABERT-2~\citep{zhou2023dnabert}.} Run with the vanilla-attention
kernel (Triton kernel disabled): the Triton path produces non-identical
representations under the same weights, so we standardise on the vanilla
implementation for all reported numbers. The model's BPE tokenizer means a
single-nucleotide variant changes the token identity at the variant
position rather than a single position within a token, which precludes masked-marginal LLR scoring at single-nt resolution.

\paragraph{Evo2-1B / 7B~\citep{brixi2026genome}.} The two variants
need different installs: Evo2-7B uses the Light path
(\texttt{vtx}~1.0.8, bf16, no \texttt{transformer\_engine}), while
Evo2-1B sets \texttt{use\_fp8\_input\_projections=True} and
requires \texttt{transformer\_engine}~$\geq$~2.3, which we build from
source in a separate environment (PyPI's TE meta-wheel pins a
\texttt{transformer\_engine\_cu12} version not published on PyPI). DMS
scoring routes through the SDK's
\texttt{score\_sequences(reduce\_method=`sum')} for whole-sequence LLR because the vortex \texttt{CharLevelTokenizer} does not expose an HF-style \texttt{\_\_call\_\_}.

\paragraph{HyenaDNA~\citep{nguyen2023hyenadna}.} Causal / autoregressive;
DMS scoring uses whole-sequence LLR rather than
masked-marginal, since the model has no mask token.
$L{=}5994$.

\paragraph{LucaOne / LucaOne-gene~\citep{he2025generalized}.} Trained with \texttt{max\_pos\_embeddings}{=}4096; we set \texttt{VEP\_\allowbreak WINDOW\_\allowbreak SIZE}{=}4094 to stay within the trained range. The shared gene--protein vocabulary is disambiguated by a \texttt{seq\_type} keyword (\texttt{`gene'}~/~\texttt{`prot'}) supplied at tokenisation time. The upstream \texttt{batch\_encode\_plus} reads from the wrong argument slot, so inputs are tokenised one record at a time through a thin adapter. Joint DNA+protein evaluation (TFIso) instantiates two LucaOne forward graphs sharing the same weights but distinct \texttt{seq\_type} tags.

\paragraph{NTv3-650M / 100M~\citep{boshar2025foundational}.} The U-Net architecture
requires sequence length divisible by $128$ (seven downsample stages); we
pad with the $N$-token (a real token in NTv3's alphabet) rather than
\texttt{[PAD]}. VEP uses $L{=}4096$ (next multiple of 128 below the trained
context), no leading or trailing special tokens, and we walk
\texttt{embed\_dim} (rather than \texttt{hidden\_size}) to retrieve the
representation width.

\paragraph{Omni-DNA-116M / 300M~\citep{li2025omni}.} Hard 250 BPE-token
context cap. For VEP, we set $L{=}1200$~bp (about 244 BPE tokens, leaving
room for special tokens). Three of the five DMS sequences exceed the
250-token budget, so DMS scoring uses a \emph{mutation-centred} 246-token
window: we locate the BPE token covering the variant nucleotide, slide a
246-token window centred on it (clamped if it would overshoot the
sequence), and compute LLR on that window only. Wildtype log-likelihood is
cached per (sequence, mutation-position) pair, giving roughly $3\times$ WT
re-use under DMS's average of 3 alleles per site. The same windowing
logic is applied in the embedding-suite path so that all DMS metrics for
Omni-DNA are computed on identical token windows.

\paragraph{Orthrus-4track~\citep{fradkin2026orthrus}.} A contrastive Mamba
SSM trained with one-hot inputs (no learned token embedding). We wrote a
tokenizer adapter that maps single nucleotides to integer ids and a model
wrapper that converts those ids to a one-hot tensor inside the forward
pass. A small \texttt{Block} class shim bridges the \texttt{mamba\_ssm}
1.x API the released checkpoint expects to the 2.x runtime in our
environment. The model is encoder-only with no language-model head, so
DMS-LLR and VEP-LLR are undefined; only embedding-based metrics are
reported.

\paragraph{JEPA-DNABERT-2~\citep{larey2026jepa}.}
DNABERT-2-117M~\citep{zhou2023dnabert} continually pretrained with a joint-embedding predictive objective, token-level MLM, and VICReg regularisation. We load the backbone via the standard path and apply the JEPA checkpoint weights. Vanilla attention is used (Triton kernel disabled) to match the base DNABERT-2 configuration, and the checkpoint includes the MLM head, which was updated during JEPA training. VEP uses $L{=}5994$.

\paragraph{JEPA-HyenaDNA-tiny~\citep{larey2026jepa}.}
HyenaDNA-tiny-16k-d128~\citep{nguyen2023hyenadna} (0.44M parameters, 16~kb context) continually pretrained with the JEPA objective. The architecture is autoregressive, so DMS scoring uses whole-sequence LLR rather than masked-marginal scoring. We benchmark the corresponding base model (\textsc{HyenaDNA-tiny}) alongside; this is distinct from the larger \textsc{HyenaDNA} variant (large-1m, 6.6M parameters) reported elsewhere. VEP uses $L{=}5994$.

\paragraph{JEPA-NTv3-100M~\citep{larey2026jepa}.}
NTv3-100M (\texttt{NTv3\_100M\_pre})~\citep{boshar2025foundational} continually pretrained with the JEPA objective. The U-Net architecture requires sequence lengths divisible by 128, so we pad with the $N$-token (a real token in NTv3's vocabulary) rather than \texttt{[PAD]}. We benchmark the corresponding base model (\textsc{NTv3-100M}) alongside; this is distinct from the larger \textsc{NTv3-650M} variant reported elsewhere. VEP uses $L{=}4096$.

\paragraph{Carbon-3B.}
A Llama-style autoregressive DNA LM, \texttt{HuggingFaceBio/Carbon-3B}~\citep{allal2026carbon}, with a hybrid tokenizer comprising a complete 6-mer vocabulary (4096 DNA 6-mers) plus Qwen3-BPE for English, switched by a \texttt{<dna>} tag that we prepend automatically. Tiling is fixed and non-overlapping ($1 + \lceil L/6 \rceil$ tokens), so a SNV flips exactly one token and whole-sequence LLR is well-defined. The model has no mask token, so DMS routes to whole-sequence LLR and VEP to the embedding suite, following the same handling as Evo2. Loaded in bf16; VEP uses $L{=}5994$.

\paragraph{ESM-1b / ESM-1v~\citep{rives2021}.} 1024-position context cap
(1022 usable amino acids). Of the five DMS studies, only Garvie 2021
(PDE3A, 1141~aa) overflows; we tail-crop to the last 1022 aa. Because
every PDE3A variant in this dataset sits at residue $\geq 668$, tail-cropping
is equivalent to a variant-centred window for this study while preserving
per-WT amortisation.

\paragraph{ESM-C-300M / 600M.} Loaded through the official
EvolutionaryScale SDK (\texttt{esm}~3.2.1) in a dedicated environment
(\texttt{torch}~2.3.1). A thin HF-compatible adapter wraps the SDK's
forward pass so that the rest of the evaluation pipeline (logits,
\texttt{last\_hidden\_state}, hidden-states tuple) is shape-compatible
with the encoder-MLM path used for all other protein models.

\paragraph{ProtT5.} An encoder-only T5 trained with span-corruption
rather than single-token masked language modelling, so masked-marginal
DMS-LLR is undefined; we report the embedding-based DMS metrics only.

\subsection{DMS Scoring}

DMS embedding-based regression uses 5-fold cross-validation with
\texttt{GroupKFold} grouped by amino-acid-level mutation identity. The
codon-level JSONL rows in four of the five DMS studies have $1.66\times$ to
$3.02\times$ codon expansion (synonymous codons mapping to the same
mutation), and a random 5-fold split would leaks codon twins between train and
test, inflating Spearman by $0.05$--$0.27$ per study. The probe is a
\texttt{StandardScaler} followed by Ridge regression with $\alpha{=}10^{-3}$
and the Cholesky solver.

\subsection{Variant Effect Prediction}

\paragraph{Per-model context window.} The default benchmark window is $L=5,994$~bp (a 6-mer-aligned $\pm$2,997~nt window centred on the variant), used for DNABERT-2, JEPA-DNABERT-2, HyenaDNA, HyenaDNA-tiny, JEPA-HyenaDNA-tiny, Caduceus-PH/PS, NTv2-500M, Orthrus, Evo2, Carbon-3B, and AlphaGenome. Models with stricter caps use:

\begin{table}[htbp]
\centering
\caption{Per-model VEP context-window deviations from the default.}
\label{tab:vep_window}
\begin{tabular}{lcl}
    \toprule
    Model & VEP window ($L$) & Reason \\
    \midrule
    LucaOne / LucaOne-gene & 4,094 & \texttt{max\_pos\_embeddings}{=}4096 \\
    AIDO.DNA-300M          & 3,998 & \texttt{max\_pos\_embeddings}{=}4000 \\
    NTv3-650M / 100M, JEPA-NTv3-100M & 4,096 & U-Net 7-downsample (must be $\div 128$) \\
    Omni-DNA-116M / 300M   & 1,200 & 250 BPE-token context cap \\
\bottomrule
\end{tabular}
\end{table}

\paragraph{Strand averaging.} VEP scores are computed independently on the forward strand and the reverse complement and averaged. Both strands carry identical biological information at any chromosomal location, and for autoregressive models the reverse-complement pass also recovers ``downstream-context'' information that the causal forward pass lacks. 

\paragraph{Linear-probe configuration.} The two vector-valued metrics (\texttt{CLS}, \texttt{Mul}) are evaluated through leave-one-chromosome-out L2-regularised logistic regression (C=1, standardized features, balanced class weights, Newton-Cholesky solver) and reported as AUPRC. The two scalar metrics (\texttt{cosine}, $\ell_2$) are reported as direct AUPRC of the metric against the binary pathogenicity label.

\subsection{TFLink and GTEx Supervised Fine-Tuning}

Both tasks use the same fine-tuning regime as the \IsoAlign rows:
LoRA with $r{=}4$ on the DNA encoder's query and value
attention projections (about 1.4M trainable adapter parameters), a frozen
protein encoder, and the standardised MLP head described in
Appendix~\ref{app:ablation}. The LucaOne baseline uses the same LoRA
configuration applied to its query and value projections. Because LucaOne
is a single transformer that handles both nucleotide and amino-acid
inputs, we instantiate two separate forward graphs sharing the same
backbone weights but with distinct \texttt{seq\_type} tags
(\texttt{`gene'} for DNA, \texttt{`prot'} for protein) so that the
DNA-side LoRA adapters do not modify the protein representation.

\section{Sequence Ablation Details}
  \label{app:ablation}

  \subsection{Experimental Setup}

  We train DP-JEPA variants on systematically modified DNA-protein pairs to isolate the contribution of different sequence features. Each ablation uses identical architecture (NTv2-500M DNA encoder + ESM2-150M
  protein encoder) and training configuration, varying only the input sequences. Models are trained for 50 epochs with early stopping based on validation loss. We evaluate bidirectional retrieval using
  DNA-to-Protein (D2P) and Protein-to-DNA (P2D) Mean Reciprocal Rank (MRR) on a held-out validation set.

\iftoggle{arxiv}{\newpage}{}

  \subsection{Ablation Categories}

  \paragraph{Baselines.}
  \begin{itemize}[leftmargin=*,nosep]
      \item \textbf{Feature-Aligned (CDS):} Coding sequences with strict 3:1 codon-to-amino-acid alignment preserving exon boundaries.
      \item \textbf{Canonical CDS:} Standard CDS extraction without explicit feature alignment.
      \item \textbf{Transcript-Specific:} Full spliced transcripts including UTRs with CDS boundaries marked.
  \end{itemize}

  \paragraph{Coding Ablations.}
  \begin{itemize}[leftmargin=*,nosep]
      \item \textbf{Frame Shift (+1):} Shift reading frame by 1 nucleotide, disrupting codon boundaries and producing an entirely different amino acid sequence while preserving nucleotide composition.
      \item \textbf{Codon Shuffle:} Replace each codon with a randomly selected synonymous codon encoding the same amino acid, preserving the protein sequence while altering codon usage patterns.
      \item \textbf{Splice Junctions Only:} Intronic sequences flanking annotated splice sites ($\pm$50 bp), containing only splice donor and acceptor motifs.
  \end{itemize}

  \paragraph{Non-Coding Ablations.}
  \begin{itemize}[leftmargin=*,nosep]
      \item \textbf{5$'$ UTR Only:} 5$'$ untranslated regions only, excluding all coding sequence.
      \item \textbf{3$'$ UTR Only:} 3$'$ untranslated regions only, excluding all coding sequence.
      \item \textbf{UTRs Combined:} Concatenated 5$'$ and 3$'$ UTRs without CDS.
  \end{itemize}

  \paragraph{Genomic Context.}
  \begin{itemize}[leftmargin=*,nosep]
      \item \textbf{Promoter + mRNA:} 1 kb upstream promoter region concatenated with the transcript.
      \item \textbf{Promoter + mRNA + Splice:} Promoter, transcript, and explicit splice junction annotations.
      \item \textbf{Gene-Level Non-Coding:} Distal regulatory sequences only, excluding all coding and proximal elements.
  \end{itemize}

  \paragraph{Negative Sampling Strategies.}
  \begin{itemize}[leftmargin=*,nosep]
      \item \textbf{Isoform Swap:} Pair each DNA sequence with a different protein isoform from the same gene.
      \item \textbf{Homolog Swap:} Pair DNA with a protein from a homologous gene family.
  \end{itemize}

  \paragraph{Negative Controls.}
  \begin{itemize}[leftmargin=*,nosep]
      \item \textbf{Shuffled DNA:} Randomly permute nucleotides within each biological DNA sequence, destroying all sequence structure while preserving nucleotide composition.
      \item \textbf{Random DNA:} Pair proteins with randomly generated DNA sequences matching overall nucleotide frequencies.
  \end{itemize}

\section{Compute Resources}                    
  \label{app:compute}     
All experiments were conducted on an internal SLURM-managed GPU cluster.        
     
  \paragraph{Pretraining.} Each training job used 4$\times$ NVIDIA H100 GPUs (80GB) with DeepSpeed ZeRO Stage 2 and bf16 mixed precision. Table~\ref{tab:training_config} summarizes the training             
  hyperparameters for each objective, and Table~\ref{tab:compute_summary} reports the total compute used. The reported totals include hyperparameter searches and ablation studies; average run duration was  
  $\sim$8 hours ($\sim$32 GPU-hours).
          
  \paragraph{Supervised Fine-tuning.} SFT experiments used a single NVIDIA A40 GPU (48GB). GTEx fine-tuning on 10\% downsampled data ($\sim$35k training samples) took $\sim$2 hours per epoch. We used LoRA
  adapters with learning rate 1e-4, batch size 16, and early stopping (patience 4). Training typically converged within 5--10 epochs.

  \paragraph{Evaluation.} Zero-shot evaluations used a single NVIDIA A100/H100 GPU (80GB) with batch size 256. We evaluated 105 model configurations across 4 benchmark tasks (DMS, GTEx, TF-Iso). Each model evaluation completed within $\sim$1--2 hours depending on task complexity.

  \begin{table}[htbp]
      \centering
      \caption{Training hyperparameters by objective.}
      \label{tab:training_config}
      \small
      \begin{tabular}{lcccc}
      \toprule
      \textbf{Parameter} & \textbf{DP-JEPA} & \textbf{IM-JEPA} & \textbf{Contrastive} & \textbf{MLM} \\
      \midrule
      Batch size (per GPU) & 16--32 & 8 & 16 & 16 \\
      Gradient accumulation & 4 & 8 & 4 & 4 \\
      Effective batch size & 256 & 256 & 256 & 256 \\
      Learning rate & 1e-5 & 1e-5 & 1e-5 & 1e-4 \\
      Max epochs & 150 & 150 & 150 & 150 \\
      \bottomrule
      \end{tabular}
  \end{table}

  \begin{table}[htbp]
      \centering
      \caption{Total compute used for pretraining experiments.}
      \label{tab:compute_summary}
      \begin{tabular}{lccc}
      \toprule
      \textbf{Objective} & \textbf{Runs} & \textbf{Wall-Clock (h)} & \textbf{GPU-Hours} \\
      \midrule
      DP-JEPA & 34 & 274 & 1,097 \\
      IM-JEPA & 12 & 136 & 543 \\
      Contrastive & 48 & 368 & 1,474 \\
      MLM & 11 & 63 & 254 \\
      \midrule
      \textbf{Total} & \textbf{71} & \textbf{568} & \textbf{2,270} \\
      \bottomrule
      \end{tabular}
  \end{table}

\end{document}